\documentclass{article}
\usepackage[numbers]{natbib}

\usepackage[preprint]{neurips_2021}

\usepackage[utf8]{inputenc} 
\usepackage[T1]{fontenc}    
\usepackage{hyperref}       
\usepackage{url}            
\usepackage{booktabs}       
\usepackage{amsfonts}       
\usepackage{nicefrac}       
\usepackage{microtype}      
\usepackage{xcolor}         
\usepackage[ruled,vlined]{algorithm2e}
\usepackage{graphicx}
\usepackage{multirow}

\title{Unsupervised Model Drift Estimation with \\
Batch Normalization Statistics for \\ Dataset Shift Detection and Model Selection}

\newcommand*\samethanks[1][\value{footnote}]{\footnotemark[#1]}

\author{
  Wonju~Lee\thanks{equal contribution} \\
  Intel Corp.\\
  Seoul, South Korea 06134 \\
  \texttt{wonju.lee@intel.com} \\
  \And
  Seok-Yong~Byun\samethanks \\
  Intel Corp. \\
  Seoul, South Korea 06134 \\
  \texttt{mark.byun@intel.com} \\
  \AND
  Jooeun~Kim \\
  Seoul National University \\
  Seoul, South Korea 08826 \\
  \texttt{kje980714@snu.ac.kr} \\
  \And
  Minje~Park \\
  Intel Corp. \\
  Seoul, South Korea 06134 \\
  \texttt{minje.park@intel.com} \\
  \And
  Kirill~Chechil \\
  Intel Corp. \\
  Nizhny Novgorod, Russia 603024 \\
  \texttt{kirill.chechil@intel.com} \\
}

\begin{document}

\maketitle

\begin{abstract}
While many real-world data streams imply that they change frequently in a non-stationary way, most of deep learning methods optimize neural networks on training data, and this leads to severe performance degradation when dataset shift happens. However, it is less possible to annotate or inspect newly streamed data by humans, and thus it is desired to measure model drift at inference time in an unsupervised manner. In this paper, we propose a novel method of model drift estimation by exploiting statistics of batch normalization layer on unlabeled test data. To remedy possible sampling error of streamed input data, we adopt low-rank approximation to each representational layer. We show the effectiveness of our method not only on dataset shift detection but also on model selection when there are multiple candidate models among model zoo or training trajectories in an unsupervised way. We further demonstrate the consistency of our method by comparing model drift scores between different network architectures.
\end{abstract}

\section{Introduction}
With a Bayes classifier, an optimized model $\hat{\theta}$ with maximum a posterior is given by $\hat{\theta} = \arg \max_{\theta} P(X,Y|\theta)P(\theta)$ with an input $X$ and a target $Y$. The {\it dataset shift} problem, i.e., the joint probability of input and output in test data differs from that in train data, severely degrades the performance of deep neural networks during post-deployment phase~\cite{Amodei2016}, while many real-world data streams imply that they change frequently in a non-stationary way~\cite{Gama2010}. However, it is less possible to annotate or steadily inspect newly streamed data by humans, and hence it is desired to measure the model drift from the target dataset in an unsupervised manner. The dataset shift problem is reformulated into predicting test error without annotation and it has not been fully studied yet to measure the discrepancy of target dataset depending only on model parameters.

In transfer learning, predicting test error is also concerned to choose the model having the most similar inductive biases for target dataset in terms of {\it model selection}.
Thanks to the prevalence of open datasets and model zoos, it has become more important to select a model having the best initialization for fine-tuning. Furthermore, to minimize covariate shift, it is desired to choose the most similar checkpoint among training trajectories. Since computing the marginal likelihood for DNN is typically intractable, Bayesian inference cannot provide a solution for measuring inductive biases~\cite{Lyle2020} and variational Bayes methods are also infeasible due to their expensive computations~\cite{Blundell2015}.

\begin{figure} \label{fig:recovery}
	\centering
	\includegraphics[scale=0.5]{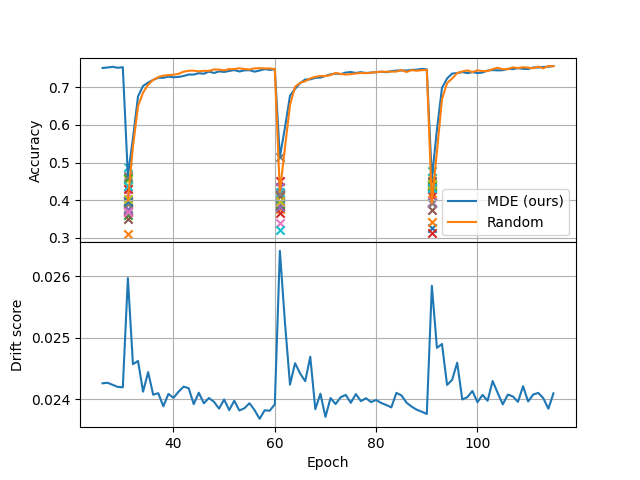}
	\vspace{-2mm}
	\caption{Automatic model selection with model drift estimation (MDE) for dataset shift detection where the dataset shift occurs at every 30 epochs. Markers indicate the accuracies of 20 different model candidates, which are initialized (or trained) with different datasets. Top: Accuracy drop happens when dataset shift happens and the recovery with the model having the least model drift score from candidates. Bottom: Estimated model drift score.}
	\vspace{-3mm}
\end{figure}

Inspired by the operation of batch normalization (BN) layer~\cite{Ioeff2015} which accumulates the statistics of data with exponential moving average during training and normalizes the input of each representational layer by these statistics during testing, we shed light on the intrinsic meaning of learnable parameters and running estimates in BN layers to measure the model drift from target datasets. Our contribution can be summarized as follows:
\begin{itemize}
\item We introduce a novel model drift estimation (MDE) method that doesn't require accessing the original train data nor annotated labels of target data. Our method exploits the capture statistics of the train data within BN and measure statistical changes while inferencing new test data. We further adopt a low-rank approximation method to each representational layer for reducing the sampling error that approximates the population statistics with sample statistics.
\item We demonstrate the effectiveness of our MDE method in dataset shift settings including both covariate shift and concept shift exist. In order to simulate the concept shift, we introduce a new label overlapping test and leverage the relationship between accuracy and drift by linear modeling. 
\item We empirically evaluate the performance of our MDE in terms of model selection for recovering from concept shift and for fine-tuning from model zoo. It is worth noting that the proposed MDE simultaneously provides dataset shift detection and model selection for recovering from dataset shift without any use of annotation.
\end{itemize}
Figure~\ref{fig:recovery} shows an example of practical uses of the MDE where the dataset shift occurs every $30$ epoch. When dataset shift happens, the drift score of MDE bounces upward and we select the model having the smallest drift from $20$ different model candidates, which are trained with different datasets, for automatic recovery from dataset shift. 



\section{Related Work}
\paragraph{Dataset Shift Detection.} 
Moreno-Torres et al.~\cite{Moreno2012} defined the dataset shifts as cases where the joint distribution of input and output differs between train and test datasets, i.e., ${\rm{P_{tr}}}\left(x,y\right) \neq {\rm{P_{te}}}\left(x, y\right)$. According to its conditional and prior probabilities, dataset shift is broadly classified into covariate shift, concept shift, and prior probability shift. Without consideration of prior probability shift for generative models~\cite{Lu2019}, we here focus on covariate shift and concept shift problems. By definition, covariate shift is occurred when
\begin{equation} \label{eq:covar}
{\rm{P_{tr}}}\left(x\right) \neq {\rm{P_{te}}}\left(x\right) \quad {\rm{and}} \quad {\rm{P_{tr}}}\left(y|x\right) = {\rm{P_{te}}}\left(y|x\right),
\end{equation}
and concept shift is occurred when
\begin{equation}\label{eq:concept}
{\rm{P_{tr}}}\left(x\right) = {\rm{P_{te}}}\left(x\right) \quad {\rm{and}}\quad {\rm{P_{tr}}}\left(y|x\right) \neq {\rm{P_{te}}}\left(y|x\right).
\end{equation}
Ovadia et al.~\cite{Ovadia2019} proposed an uncertainty quantification (UQ) benchmark for detecting covariate and concept shifts and they introduced Bayesian and non-Bayesian detection metrics, e.g., Negative log-likelihood (NLL), Brier score, and expected calibration error (ECE), by approximating the marginal likelihood. Uncertainty calibration error (UCE) has been proposed to measure mis-calibration of uncertainty that represents the difference between model error and its uncertainty~\cite{Laves2019}. Ackerman et al.~\cite{Ackerman2021} detected dataset shift by non-parametric testing the distribution of model prediction confidence. They verified the robustness of dataset shift detection from Type-I error by inserting unseen digit class in MNIST dataset. However, existing detection metrics assume a  supervised setting, and hence these are less feasible to machine learning practitioners. Moreover, they fail to detect the concept shift because of a lack of sufficient label information.

\paragraph{Model Selection}
It is required to choose the most adequate model that has similar inductive biases of model parameters to a target dataset. More specifically, transferability estimation aims to choose the model without training on the target datasets and we expect higher transferability leads to higher accuracy, especially for short training with a small training dataset. To estimate the transferability, Cui et al.~\cite{Cui2018} directly measured the domain similarity between the source and target datasets. Representation similarity analysis (RSA) computes the dissimilarity matrices of features from pretrained models and a trained model on the target dataset ~\cite{Dwivedi2019}. Nguyen et al.~\cite{Nguyen2020} proposed the log expected empirical prediction (LEEP) metric from conditional cross-entropy between the pseudo-labels from a forward-passing target data into a model and target labels. Deshpande et al.~\cite{Deshpande2021} approximate the fine-tuning dynamics by linearization of the model parameter inspired by Neural Tangent Kernel and compute the similarity from its gradient and labels of the target dataset. They further propose the baseline framework for model selection for fine-tuning. Lyle et al.~\cite{Lyle2020} illustrated the relationship between training speed and marginal likelihood from Bayesian inference.
Otherwise, Ueno $\&$ Kondo~\cite{Ueno2021} introduce several metrics for model selection including unsupervised feature map sparsity analysis (FSA). The above supervised approaches include target labels or even training processes and it is still an open question to measure transferability from the model parameters itself.


%

\section{Model Drift Estimation} \label{sec:algo}
With an assumption that BN layers capture the statistics of train data on both learnable parameters and running estimates, we exploit such statistics of BN layers to estimate the discrepancy of the train data from target data in a context of the trained model. The proposed MDE effectively estimates the test error with a single forward passing the unlabeled data. In this section, we first describe the operation of BN layers, and then describe our MDE method in details.  To increase the reliability, we further adopt a low-rank approximation. The detailed algorithm is illustrated in Algorithm~\ref{algo:algo}.

\subsection{Batch Normalization}
It is known that BN operation smooths the optimization landscape in the parameter space during training DNN by normalizing each input feature within a mini-batch to have a zero-mean and unit-variance~\cite{Santurkar2018}. And hence, especially for visual tasks, most neural network architectures include the BN layer due to its gain of stability and convergence speed. Let us define the input feature vector of mini-batch $n$ as $x^{n} = \left[ x_{1}^{n}, x_{2}^{n} \dots, x_{C}^{n} \right]$ with the number $C$ of channels.  BN layer normalizes the input feature vector by its stored running estimates, i.e., running mean $\mu$ and running standard deviation $\sigma$, and BN layer outputs the normalized signal by multiplying $\gamma$ and adding $\beta$ as
\begin{equation} \label{eq:bn}
y_{c}^{n} = \gamma_{c} \frac{x_{c}^{n}-\mu_{c}}{\sigma_{c}} + \beta_{c}.
\end{equation}
We denote that $\gamma_{c}$ and $\beta_{c}$ are learnable parameters and running estimates $\mu_{c}$ and $\sigma_{c}$ are calculated following exponential moving average during a training phase.
That is, we infer that the statistics of source data are implicitly included in the running estimates and we can measure the model discrepancy by comparing statistics of source data in the model and statistics of target data without label information.

\subsection{Model Drift Score Based on Batch Normalization Statistics}
Model drift is usually coming from the distributional shift between the source and target datasets, i.e., the model is trained on the source, but the model is deployed on the target. Since it is in general infeasible to suppose the accessibility to source data in artificial intelligence practices, we propose a novel drift estimation methodology with no need of accessing source data as well as requiring annotated labels of target data. Reminding the fact that the running estimates accumulate the statistics of source data and are utilized for inference time, we find that BN layers implicitly can  aid estimating the discrepancy of target data from source data. Specifically, with BN layers, exponential moving averaged statistics $\mu_{c}$ and $\sigma_{c}$ of source can be compared with statistics $\bar{\mu}_{c}$ and $\bar{\sigma}_{c}$ of target, where $\bar{\mu}_{c}^{n}$ and $\bar{\sigma}_{c}^{n}$ are mean and standard deviation of mini-batch $x^{n}$, respectively. By normalizing the output feature vectors $y_{c}^{n}$ of mini-batch $n$ with learnable parameters $\gamma_{c}$ and $\beta_{c}$, the proposed drift score of BN layer $l$ can be obtained by
\begin{equation} \label{eq:sim}
d^{(l)} = \frac{1}{NC}\sum_{n=1}^{N}\sum_{c=1}^{C}{\rm{Dist}}\left( \frac{x_{c}^{n, (l)}-\bar{\mu}_{c}^{n,(l)}}{\bar{\sigma}_{c}^{n,(l)}}, \frac{y_{c}^{n, (l)}-\beta_{c}^{(l)}}{\gamma_{c}^{(l)}} \right),
\end{equation}
where ${\rm{Dist}}(a,b)$ is a distance metric between vectors $a$ and $b$. From the Equation~\ref{eq:sim}, the right element of distance metric contains the information of the source in the BN layer and the left element that contains the information of the target, and hence we can implicitly compute the discrepancy  of the source dataset from target dataset using only model parameters. Finally, overall model discrepancy can be calculated by taking average of all layers as
\begin{equation}\label{eq:drift}
D = \frac{1}{L}\sum_{l=1}^{L}w^{(l)}d^{(l)},
\end{equation}
where weight $w^{(l)}\in \left[0,1\right]$ indicates the relative importance of BN layer $l$ compared to others. The weights can be set proportional to magnitude of gradient during training phase~\cite{Singh2015} or ratio the L2-norm of weight and gradient~\cite{You2017}.
In this paper, we simply set $w^{(l)}$ to $1$ for all $l$.

We suppose that input feature vector at layer $l$ is independent and identically distributed (i.i.d.) random variable following Gaussian distribution with mean $\bar{\mu}_{c}^{(l)}$ and standard deviation $\bar{\sigma}_{c}^{(l)}$. Then the left term of (\ref{eq:sim}) follows Gaussian distribution with zero-mean unit-variance, while the right term of (\ref{eq:sim}) is as follow.
\begin{equation}
\frac{y_{c}^{n, (l)}-\beta^{(l)}}{\gamma^{(l)}} \sim \mathcal{N}\left(\frac{\bar{\mu}_{c}^{n, (l)}-{\mu}_{c}^{(l)}}{\sigma_{c}^{(l)}},\left(\frac{\bar{\sigma}_{c}^{n, (l)}}{\sigma_{c}^{(l)}}\right)^{2}\right).
\end{equation}
Drift score in (\ref{eq:sim}) can be approximated into compare the distance between two random variables following Gaussian distribution, and the distance metric is simplified to the Wasserstein distance as
\begin{equation} \label{eq:simKL}
d^{(l)} \approx  \frac{1}{NC}\sum_{n=1}^{N} \sum_{c=1}^{C}\frac{(\bar{\mu}_{c}^{n, (l)}-\mu_{c}^{(l)})^{2}+(\bar{\sigma}_{c}^{n, (l)}-\sigma_{c}^{(l)})^{2}}{(\sigma_{c}^{(l)})^{2}}.
\end{equation}
This can be also replaced by Kullback-Leibler (KL) divergence or statistical z-test based distance.
In the information-theoretic viewpoint, difference between expected risks of hypothesis $h$ on the target and on the source is bounded as $r_{{\rm{T}}}(h) - r_{{\rm{S}}}(h) \leq \sqrt{D_{{\rm{KL}}}(p_{\rm{S}}, p_{\rm{T}})/2}$
with KL divergence $D_{{\rm{KL}}}$ of source distribution $p_{\rm{S}}$ and target distribution $p_{\rm{T}}$ from Pinsker's inequality~\cite{Ben-David2010, Ishii2021}.
This supports that measuring model drift from the target can be naturally extended to estimate the test error of the hypothesis on the target. 

Without the assumption that input feature vector is i.i.d. Gaussian random variable, the drift score (\ref{eq:sim}) of BN layer $l$ can be computed with conventional distance metrics such as cosine distance ${\rm{CosDist}}(a,b)=(1-a\cdot b / ||a|| ||b||) / 2$, which is bounded in $\left[0,1\right]$. 
By noting that the relative scale of the drift score is important for both shift detection and model selection, it is required to change the distance metric that does not intrude the rank order and we later select the cosine distance without making i.i.d. Gaussian assumption. Specifically, the drift score is expected to have a discriminating value to that of the identical case in shift detection problems and this should be also separate the quality of models for transferring in model selection problems.


Since our MDE method supposes that the running estimates in BN layers implicitly represent the true population statistics of source data, the performance of the MDE is highly affected by how well the running estimates approximate the true statistics. Recently, Yan et al.~\cite{Yan2020} explains the properties of running estimates in both forward and backward propagation and proves that running estimates at iteration $t$
\begin{equation} \label{eq:ema}
\mu_{c,t} = \hat{\mu}_{c}+\epsilon \quad {\rm{and}} \quad \sigma_{c,t}^2 = \frac{\left(1-\alpha^{2t}\right)\left(1-\alpha\right)}{1+\alpha}\hat{\sigma}_{c}^{2}+\epsilon
\end{equation}
holds when batch statistics weakly converges to the true batch statistics as training finishes. Here, $\hat{\mu}$ and $\hat{\sigma}$ are true population mean and standard deviation, $\alpha$ is momentum for updating the running estimates from exponential moving average and $\epsilon$ is an extremely tiny unbiased substitute of batch statistics. Consequently, the running mean is not injured from the true population mean and the smaller $\alpha$ leads the less contamination of the running standard deviation.


\subsection{Low-Rank Approximation of Representation Layer}
In this subsection, we further apply low-rank approximation to achieve dimensionality reduction of each representational layer output. The low-rank approximation preserves the invariant subspace that captures the essential information of input features, while this rejects redundancies. We adopt truncated singular value decomposition (SVD) to input feature vectors $x^{n,(l)}\in \mathbb{R}^{C \times H \times W}$ of each $l$ BN layer. We first reformulate $x^{n,(l)}$ to be $\tilde{x}^{n,(l)} \in \mathbb{R}^{C \times HW}$ and apply SVD as
\begin{equation}
\tilde{x}^{n,(l)} = UDV^{T}
\end{equation}
for $n=1, \cdots, N$, where the orthogonal matrix $U\in \mathbb{R}^{C\times C}$ contains columns of singular vectors in the output space that represent independent directions of output variants, 
the orthogonal matrix $V\in \mathbb{R}^{HW\times HW}$ contains columns of singular vectors in the input space that represent independent directions of input variants, and $D\in \mathbb{R}^{C\times HW}$ is a diagonal matrix with ordered singular values $\sigma_{1} \geq \sigma_{2} \geq \cdots \geq \sigma_{{\rm{min}}\left(C,HW\right)}$. With truncation ratio $R_{{\rm{tr}}}$, in order to reduce the distortion from sample error, we refine the input as
\begin{equation} \label{eq:tsvd}
\hat{x}^{n,(l)} = U\hat{D}V^{T}
\end{equation}
where truncated diagonal matrix $\hat{D} = {\rm{Diag}}\left(\left[ \sigma_{1}, \sigma_{2}, \cdots, \sigma_{R_{{\rm{tr}}}{\rm{min}}\left(C,HW\right)}, 0, \cdots, 0 \right] \right)$. It is worth noting that BN operation multiplying $\gamma$ and adding $\beta$ could amplify and forward the signal including sampling error and hence it deteriorates the performance of the drift measurement. By applying low-rank approximation, we overcome this degradation and enhance the reliability of MDE, especially for the region of small performance drop. The detailed algorithm of MDE with low-rank approximation is described in Algorithm~\ref{algo:algo}.

\begin{algorithm}[H] \label{algo:algo}
\SetAlgoLined
\SetKwInOut{Input}{Input }
\SetKwInOut{Output}{Output}
\Input{Dataset $B$, number of iterations $T$}
\Output{model drift $D$}
${\bf{x}} = \{ \}$, ${\bf{y}}=\{ \}$, ${\bf{w}} = \{ \}$\;
\For{$T$ {\rm{iterations}}}{
	Sample $x$ from $B$ \;
	${\bf{w}} = {\bf{w}} \cup \left\{\gamma, \beta\right\} $ \;
	${\bf{x}} = {\bf{x}} \cup {\hat{x}}$ from (\ref{eq:tsvd}) \;
	${\bf{y}} = {\bf{y}} \cup f_{\gamma,\beta}(x)$ from (\ref{eq:bn}) \;
}
Compute drift $D$ from (\ref{eq:sim}) and (\ref{eq:drift})
\caption{MDE method with low-rank approximation}
\end{algorithm}

\section{Experimental Results} \label{sec:exp}
We perform empirical evaluations of the proposed MDE on image classification task under two types of dataset shifts, i.e., covariate shift and concept shift, and analyze the performance of MDE and that with low-rank approximation by linear modeling between accuracy and drift score in Section~\ref{sec:exp_dd}. We compare the proposed MDE with Bayeisan and non-Bayesian drift detection metrics including NLL, Brier score, and ECE~\cite{Ovadia2019}. We then verify the performance of MDE in two model selection problems, where one is selecting the model from training trajectories under concept shifts and the other is choosing the most feasible model on model zoo containing the models trained on different source datasets in Section~\ref{sec:exp_ms}. We investigate the model selection performances of FSA and random selection as unsupervised approaches. We further compare the MDE with supervised methods including DSA~\cite{Cui2018}, RSA~\cite{Dwivedi2019} and LEEP~\cite{Nguyen2020}. In this paper, we train the ImageNet pretrained ResNet-18 with Adam optimizer for $50$ epochs.
We set the batch size to $128$ and the initial learning rate to $0.001$ and decay the learning rate by a factor of $10$ at the $35$th epoch.

\subsection{Performance Analysis over Dataset Shift} \label{sec:exp_dd}
\paragraph{Covariate Shift} 
The distribution of images according to the target data is changed and it frequently occurs when sensor performance deteriorates, sensor equipment is changed, or an unintended object such as dust or water drop is steadily caught in the sensor. In order to simulate these situations with CIFAR-100 dataset, we corrupt images with various kinds of perturbations, e.g., geometric transformations, illuminance changes, noise injections, and random erasing with cut out, as done in~\cite{Ovadia2019, Krishnan2020}.

\begin{figure} 
	\centering
	\includegraphics[scale=0.30]{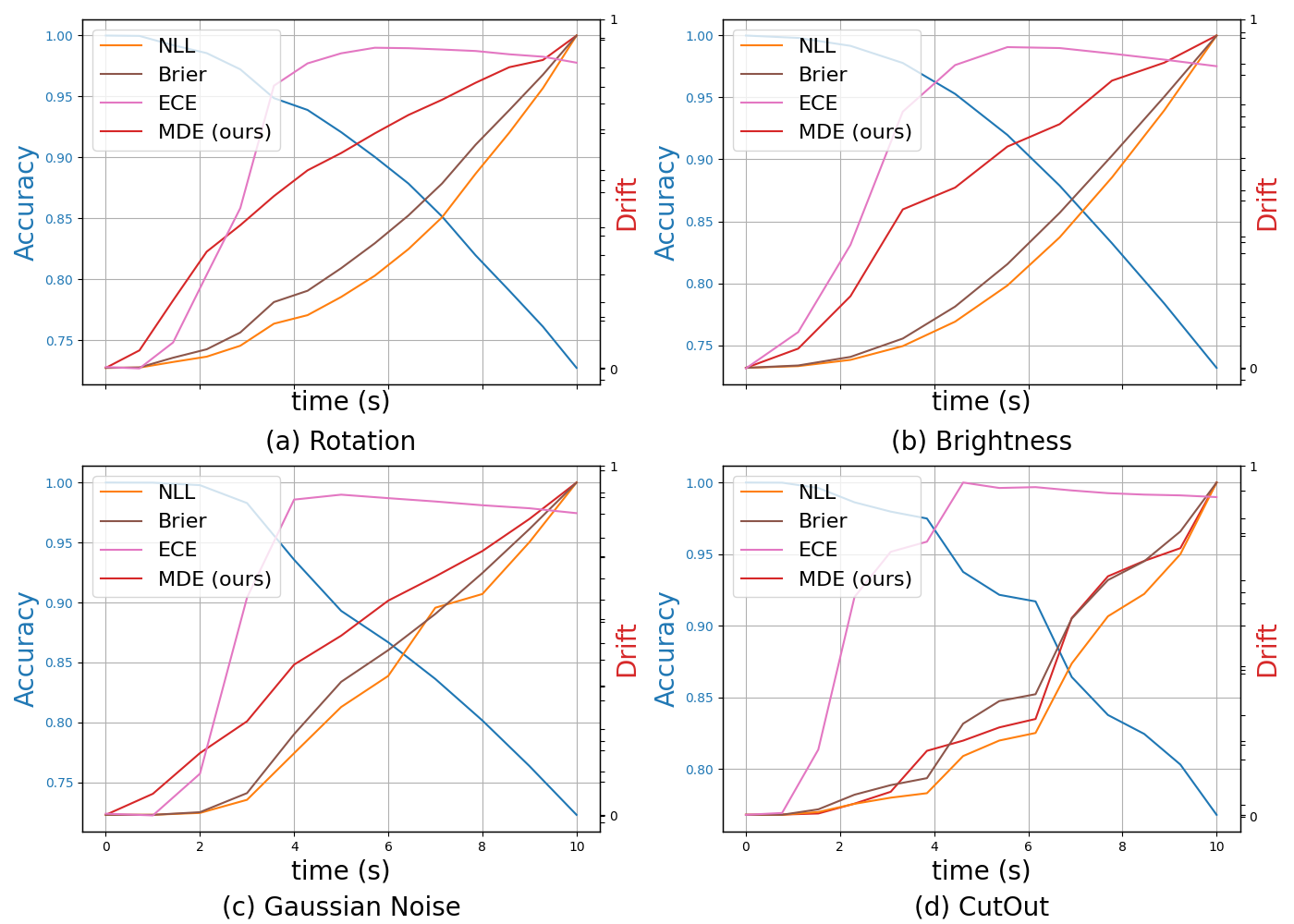}
	\vspace{-2mm}
	\caption{Changes of accuracy (red) and drift score (blue) of detection metrics as (a) rotation transforms larger, (b) brightness changes more severely, (c) variance of Gaussian noise goes larger, (d) cut-out with the number of holes and its size go larger.}
	\vspace{-2mm}
	\label{fig:covariate}
\end{figure}
In Figure~\ref{fig:covariate}, when performance drops according to perturbations, the proposed MDE identifies covariate shifts by increasing its drift score. Compared to other metrics, the Brier score is most identical to have a negative correlation with the accuracy and MDE also follows the overall tendency well, while only the MDE is an unsupervised approach. It is observed that curvatures of the MDE describe the accuracy better than ECE. 
It is further observed that the slopes of MDE with respect to distortion are different. For instance, when accuracy drops 10 \% from the identical case, the drift score increases 33$\times$ with rotational shifts, but the drift score increases less than 2$\times$ with random erasing. That is, the drift score from MDE is much sensitive to geometric transformation than random erasing and we identify that activations of the model are robust to random erasing in the perspective of interpretability of DNN.
This can be treated as a similar sense to layer-wise relevance propagation with BN layer~\cite{Guillemot2020}. From these investigations, we demonstrate the effectiveness of MDE to detect covariate shifts.

\paragraph{Concept Shift} 
Contrary to covariate shift, from (\ref{eq:concept}), the distribution of input according to the target dataset is maintained, but the distribution of target is changed during the post-deployment phase.
Ovadia et al.~\cite{Ovadia2019} suppose that concept drift occurs when the model is learned on CIFAR-10, but the model is deployed on SVHN dataset. In order to make controllability and smoothing this testing scenario, we propose a new framework, {\it{overlapping test}}, which divides the overall dataset into training and test data according to its class information by adjusting overlapping probability.
With the number $50k$ of data in CIFAR-100, when the probability of overlapping is $0.33$, the number $30k$ of training data and $30k$ of validation data are composed of having $60$ classes respectively and $20$ classes are overlapped between training and validation data. The overlapping test is an extension of experiments adding unseen class data at the testing phase as done in~\cite{Ovadia2019, Cerquitelli2019, Ackerman2021} and can elaborately depict the concept drift test with hierarchical datasets such as CIFAR-100. For instance, the model is trained to classify the maples, oaks, and palms, but the model has to classify the pine and willow in the post-deployment phase. Specifically, all the above $5$ are categorized into tree superclass in CIFAR-100, however, the model fails to classify images according to unseen classes due to the concept shift problem.

\begin{figure} 
	\centering
	\includegraphics[scale=0.5]{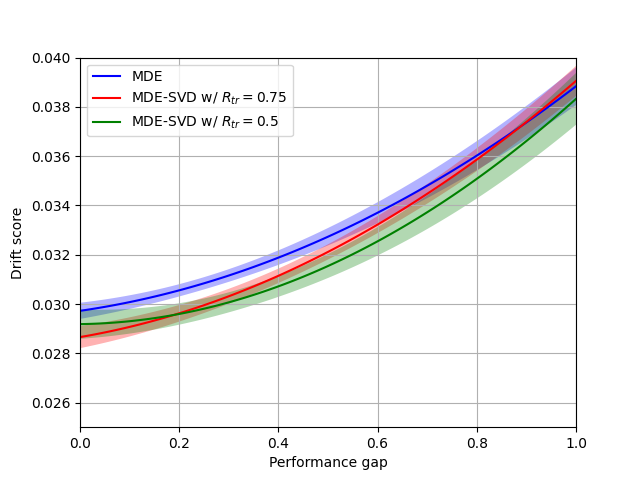}
	\vspace{-2mm}
	\caption{Drift score of the MDE with respect to performance gap between training and test data in overlapping test, where the shaded regions show $95 \%$ confidence intervals.}
	\vspace{-0mm}
	\label{fig:concept}
\end{figure}

In order to demonstrate the feasibility of the proposed MDE for estimating test error without annotated data, Figure~\ref{fig:concept} leverages the linear modeling between performance gap and drift score, where the shaded regions show $95 \%$ confidence intervals. From the results, the drift score from the MDE clearly helps to estimate test error while the MDE cannot significantly distinguish the small performance gap between the model trained on source data and target data. The MDE with low-rank approximation increases the reliability from its steepness of the slope in the regions of the small performance gap. However, we observe that the stability of the proposed MDE is degraded from its high fluctuations of those with low-rank approximation. This is because the low-rank approximation disturbs the invariant subspace containing essential information in input feature vectors. We can further investigate reliability degradation from the result of $50 \%$ of truncation in the viewpoints of reliability and stability.

\begin{table}
\caption{Performance of model selection schemes under concept shift}
\small \centering
\begin{tabular}{clccc|cccc}
\toprule
\multirow{2}{*}{c}  & \multicolumn{1}{c}{\multirow{2}{*}{\begin{tabular}[c]{@{}c@{}}Perfor-\\ mance\end{tabular}}} & \multicolumn{3}{c|}{Supervised} & \multicolumn{4}{c}{Unsupervised}   \\
& \multicolumn{1}{c}{}                                                      & DSA & RSA & LEEP & FSA & Random & MDE & MDE-SVD \\ \midrule
\multirow{3}{*}{25} 
& top-3 acc.                                            
& 0.8752    & 0.5515   & 0.7562   & 0.3683 & 0.1500 & 0.4525 & \bf{0.4530}  \\
& top-5 acc.                
& 0.9590    & 0.6623   & 0.8872   & 0.5005 & 0.2000 & 0.6120 & \bf{0.6251}  \\
& top-1 err.                                                                & 0.0123    & 0.0504   & 0.0251   & 0.0836 & 0.1312 & 0.0691 & \bf{0.0687}  \\ \midrule
\multirow{3}{*}{50} 
& top-3 acc.                                                                & 0.8854    & 0.5185   & 0.7517   & 0.3102 & 0.1500 & 0.3183 & \bf{0.3298}  \\
& top-5 acc.                                                                & 0.9629    & 0.6200   & 0.9016   & 0.4278 & 0.2000 & 0.4469 & \bf{0.4675}  \\
& top-1 err.                                                                & 0.0081    & 0.0372   & 0.0144   & 0.0578 & 0.0810 & 0.0535 & \bf{0.0505} \\ \midrule
\multirow{3}{*}{75} 
& top-3 acc.                                                                & 0.8232    & 0.4600   & 0.7461   & 0.2337 & 0.1500 & 0.2284 & \bf{0.2571}  \\
& top-5 acc.                                                                & 0.9264    & 0.5880   & 0.8757   & 0.3442 & 0.2000 & 0.3604 & \bf{0.4221}  \\
& top-1 err.                                                                & 0.0054    & 0.0200   & 0.0085   & 0.0363 & 0.0422 & 0.0332 & \bf{0.0292}  \\ \bottomrule
\end{tabular}
\vspace{-2mm}
\label{tbl:remedy}
\end{table}

\subsection{Performance Analysis over Model Selection} \label{sec:exp_ms}
\paragraph{Recovery from Concept Shift} 
Like overlapping tests, we extract the data according to randomly chosen $c$ classes per cycle, and each cycle is iterated for every 30 epochs. Then, the training datasets among cycles could be partially overlapped and hence the model parameters naturally suffer the concept shift problem. We here implement the automatic remedy from concept drift by choosing the model yielding the smallest drift score among previously trained experts which are optimized on each source data. Figure~\ref{fig:recovery} shows that the remedy model having the smallest drift score among $20$ experts is quite preservative to performance drop from concept shift with $c=50$. We then measure the top-$k$ accuracy of model selection with the $20$ experts model trained on source data of randomly chosen $c$ classes in Table~\ref{tbl:remedy}. As $c$ increases, i.e., concept drift is small, the performances of top-$k$ are degraded but the gap between the most feasible model from the MDE and the optimal selection is going to be lower. Especially for $c=75$ where the probability of overlapping between two consecutive cycles is $(75/100)^2 = 0.5625$, the model selection from the MDE losses about $2.9 \%$ of accuracy from the optimal selection, but that from the MDE gains about $1.3 \%$ from random selection and about $4.9 \%$ from the worst selection. Furthermore, the proposed MDE with the low-rank approximation is more effective for a larger $c$ due to its increase of reliability as already described in Figure~\ref{fig:concept}. From the results, it is meaningful that the model selection according to the proposed MDE effectively defends against performance degradation from concept shift without any use of annotated data.


\paragraph{Model Selection among Model Zoo} 
To demonstrate of performance of the proposed MDE in the purpose of choosing the best initialization model, we fine-tune the pretrained model on the target dataset with a short training and compare the accuracy of the model candidates. We here note that the final accuracy over longer training is less meaningful in the view of model selection because the final accuracy is highly affected by hyperparameter optimization for training strategy rather than initial model parameters. We prepare the pretrained models on various kinds of source datasets, e.g., ImageNet, CIFAR-100, VisDA2020 (Real, Sketch, Infograph, Clipart, Painting, Quickdraw). With $8$ numbers of datasets, we generate the $64$ numbers of pairs of train on source dataset $i$ and fine-tune on target dataset $j$ for $i,j \in \{1,2,\cdots,8\}$. In this work, it is emphasized that the proposed MDE does not use any information from the source dataset and concentrates only on the network parameters. In this experiment, we use Resnet-50 for fine-tuning the target dataset for $1$ epoch training to validate the best initialization model. We remind that the performance of the relative scale of the drift scores among model candidates at a given target dataset is more important for model selection performance.

\begin{figure}
	\centering
	\includegraphics[scale=0.45]{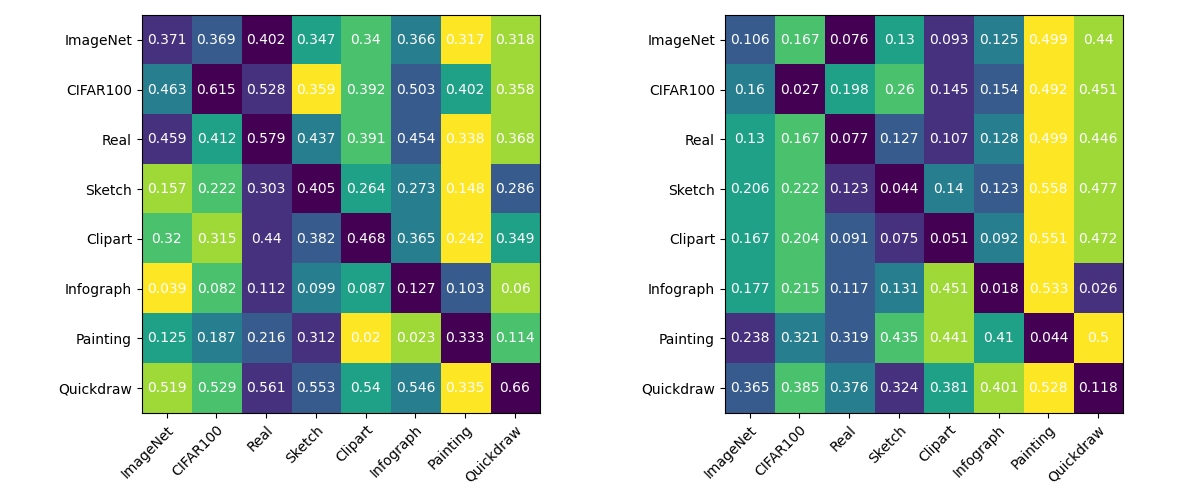}
	\vspace{-3mm}
	\caption{Heatmap shows ranking of fine-tuning accuracies and that of drift score for demonstrating the best initialization model, where x and y-axis represent the source and target datasets, respectively. Left: Accuracy of pretrained models for $1$ epoch training on target dataset, Right: Drift score of pretrained models on target dataset.}
	\vspace{-0mm}
	 \label{fig:ms}
\end{figure}


\begin{table}
\caption{Performance of model selection schemes for fine-tuning from model zoo}
\small
\centering
\begin{tabular}{lccccc}
\toprule
\multirow{2}{*}{Method} & \multicolumn{2}{c}{Source} & \multicolumn{2}{c}{Target} & \multirow{2}{*}{\begin{tabular}[c]{@{}c@{}}Spearman's\\ rank corr.\end{tabular}} \\ \cline{2-5}
                        & Data & Label  & Data & Label & \\ \midrule
DSA & O & O & O & O & 0.5580 \\ \midrule
RSA	& X  & X & O & O & 0.6041 \\ \midrule
LEEP & X & X & O & O & \bf{0.6904} \\ \midrule
FSA & X & X & O & X & 0.3869 \\ \midrule
MDE & X & X & O & X & \bf{0.6279} \\ \bottomrule
\end{tabular}
\vspace{-2mm}
\label{tbl:modelzoo}
\end{table}

Figure~\ref{fig:ms} illustrates the heatmap of ranking for accuracy and drift score per pair and the number indicates its accuracy and drift score. For a given target dataset per each column, the overall tendency of rank is well maintained. It is interesting that the Real pretrained model provides the most attractive initialization while the ImageNet pretrained model is broadly considered a favorable solution. Highlighting the performance of ranking among model zoo in Table~\ref{tbl:modelzoo}, the Spearman's rank correlation of the proposed MDE is obtained by $0.6279$ and this overwhelms the unsupervised methods where FSA and random selection exhibit $0.3869$ and $0$, respectively. The proposed MDE is quite competitive to supervised approaches for model selection while MDE didn't need any additional label information of the target dataset. Furthermore, no use of target label enables smart annotations or human in the loop training methods such as active learning.

\begin{figure}
	\centering
	\includegraphics[scale=0.5]{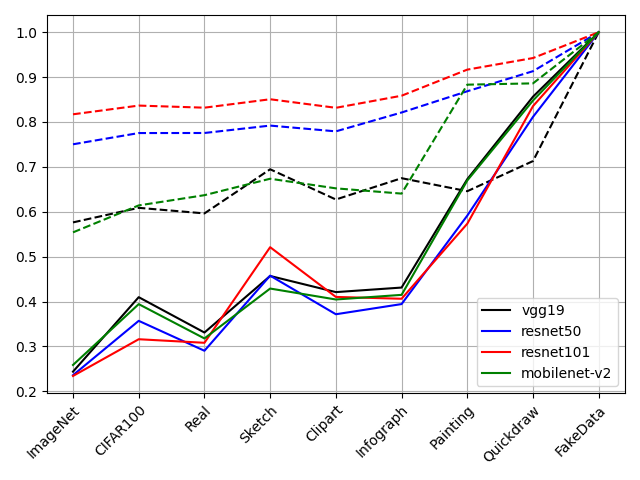}
	\vspace{-3mm}
	\caption{Figures shows drift scores of different backbone architectures with respect to target datasets, where drift scores are normalized by drift value on FakeData. Solid lines show the drift score from cosine distance (\ref{eq:sim}) and dashed lines show the approximated drift score from Wasserstein distance (\ref{eq:simKL}).}
	\vspace{-2mm}
	\label{fig:consist}
\end{figure}

\paragraph{Consistency Analysis on Various Model Architectures}
In order to see the consistency of drift score to different architectures, we adopt various backbone architectures, e.g., VGG-19, Resnet-50, Resnet-101, Mobilenet-v2 which are publicly available ImageNet pretrained models. Figure~\ref{fig:consist} shows the drift scores of different backbone architectures with respect to target datasets. We check that the drift scores on FakeData following zero-mean and unit-variance Gaussian distribution yields the maximum drift value and this could be a normalization factor between different backbones. As a result, the proposed MDE is consistent with backbone architecture and hence the proposed MDE can be useful to choose the most feasible model from the model zoo including different architectures trained on different source datasets. We next represent the performance of approximation described in (\ref{eq:simKL}) in Figure~\ref{fig:consist} as dashed lines. With the approximation, it is inconsistent to different backbones but the ranking according to the target dataset is well maintained especially for the Resnet family. By noting that the ranking or relative scale is the most important thing in both drift detection and model selection, it can be seen that the approximation provides meaningful information.

\section{Conclusion}
Detecting dataset shift and selecting the best model for transfer learning without annotation are still open but very practical problems. In this paper, we presented a novel method of estimating model drift by exploiting the statistics of batch normalization layer in an unsupervised manner. By comparing the statistics captured in pretrained BN layers and those from newly streamed input data, our method estimates the test error of the model with a single forward pass. With simple linear modeling, we demonstrated the effectiveness of our method not only for concept and covariate shift detection but also for selecting the best model among model zoo and training trajectories. Even though we provided an empirical evidence that our method can be applied to diverse network architectures having BN layers, it's still open to extend our idea to different normalization layers and architectures.

\bibliographystyle{unsrt}
\bibliography{ref}

\end{document}